# A weakly-supervised deep learning model for fast localisation and delineation of the skeleton, internal organs, and spinal canal on Whole-Body Diffusion-Weighted MRI (WB-DWI)

*Antonio Candito[1], Alina Dragan[1,2], Richard Holbrey[1], Ana Ribeiro[2], Ricardo Donners[3], Christina Messiou[1,2], Nina Tunariu[1,2], Dow-Mu Koh[1,2], and Matthew D Blackledge[1]*

[1]The Institute of Cancer Research, London, United Kingdom

[2]The Royal Marsden NHS Foundation Trust, London, United Kingdom

[3]University Hospital Basel, Basel, Switzerland

Corresponding Author: Matthew D Blackledge
The Institute of Cancer Research, London (UK)
Matthew.Blackledge@icr.ac.uk

## Abstract:

**Background**: Apparent Diffusion Coefficient (ADC) values and Total Diffusion Volume (TDV) from Whole-body diffusion-weighted MRI (WB-DWI) are recognised cancer imaging biomarkers. However, manual disease delineation for ADC and TDV measurements is unfeasible in clinical practice, demanding automation. As a first step, we propose an algorithm to generate fast and reproducible probability maps of the skeleton, adjacent internal organs (liver, spleen, urinary bladder, and kidneys), and spinal canal.

**Methods**: We developed an automated deep-learning pipeline based on a 3D patch-based Residual U-Net architecture that localises and delineates these anatomical structures on WB-DWI. The algorithm was trained using "soft-labels" (non-binary segmentations) derived from a computationally intensive atlas-based approach. For training and validation, we employed a multi-centre WB-DWI dataset comprising 532 scans from patients with Advanced Prostate Cancer (APC) or Multiple Myeloma (MM), with testing on 45 patients.

**Results**: Our weakly-supervised deep learning model achieved an average dice score of 0.67 for whole skeletal delineation, 0.76 when excluding ribcage, 0.83 for internal organs, and 0.86 for spinal canal, with average surface distances below 3mm. Relative median ADC differences between automated and manual full-body delineations were below 10%. The model was 12x faster than the atlas-based registration algorithm (25sec vs. 5min). Two experienced radiologists rated the model's outputs as either "good" or "excellent" on test scans, with inter-reader agreement from fair to substantial (Gwet's AC1=0.27-0.72).

**Conclusion**: The model offers fast, reproducible probability maps for localising and delineating body regions on WB-DWI, potentially enabling non-invasive imaging biomarkers quantification to support disease staging and treatment response assessment.

## 1. Background

Whole-body diffusion-weighted MRI (WB-DWI) enables detection, staging, and response assessment of systemic malignant disease in patients diagnosed with Advanced Prostate Cancer (APC) or Multiple Myeloma (MM) [1]–[3]. Bone lesions exhibit high DWI signal intensity due to restricted water mobility within tumour-infiltrated bone marrow [4], which can be quantified using the Apparent Diffusion Coefficient (ADC) derived from images acquired at multiple b-values (typically 0-50 and 900-1000 s/mm$^2$) [5], [6]. ADC inversely correlates with tissue cellularity and serves as a potent biomarker of response [7], [8]. Additionally, the Total Diffusion Volume (TDV, in millilitres) of delineated bone disease on WB-DWI reflects estimated tumour burden [9], [10].

However, radiologists need to manually delineate bone lesions throughout the skeleton to measure ADC and TDV. This process is cumbersome and time-consuming, often requiring an hour or more, depending on disease burden [11]. As a result, manual delineation is not feasible for routine reporting workflows. There is a growing clinical demand for software solutions that can automate this task, assisting radiologists in quantifying bone disease in patients diagnosed with APC or MM [12]. We hypothesise that a logical first step toward automation is developing an algorithm to localise and delineate the whole skeleton and surrounding anatomy from WB-DWI scans. This would isolate bone from adjacent anatomical structures and/or non-skeletal regions that exhibit hyperintense signals on b900 images. Similar two-step pipelines have been developed for delineating lytic bone lesions on whole-body CT [13] and focal bone marrow lesions on MRI of the left pelvis only [14].

Atlas-based registration algorithms and deep learning models have been proposed for delineating bone marrow from MRI sequences with higher resolution and signal-to-noise ratio (SNR) than WB-DWI. Ceranka et al. [15] developed a multi-atlas delineation algorithm for T1-weighted (T1w) MRI in APC patients, achieving a dice score of 0.88 between manual and automated masks. Lavdas et al. [16] applied atlas-based registration to T2-weighted (T2w) MRI in healthy volunteers, reporting a dice score of 0.71. However, the accuracy of atlas-derived delineations may be affected by the characteristics of the atlas cohort, including the extent of disease spread across patients and how experts defined the ground truth, as well as the initial spatial alignment between whole-body target and atlas image [17], [18]. The computational time required for registrations may limits clinical deployment.

Deep learning models aim to address these limitations. A nnU-Net trained on coronal T1w MRI for bone marrow delineation in myeloma patients achieved a dice score of 0.88 [19], [20], while D'Antonoli et al. [21] released the open-source *TotalSegmentator-MRI* model for multi-organ delineation from multiple MRI sequences. However, developing such models requires high-quality annotations across high-volume WB-DWI datasets [22], an impracticable task due to time and resource constraints commonly faced in radiology departments. Furthermore, automated delineation of the bone marrow space from WB-DWI scans remains a complex task due to inter-patient heterogeneity, the varied appearance and distribution of disease, and the presence of imaging artifacts due to low SNR or magnetic susceptibility effects [23], [24]. For these reasons, we explored a weakly-supervised training strategy using atlas-derived "soft-labels" (non-binary segmentations), which may provide a scalable and probabilistic representation of anatomical structures while accounting for uncertainty in regions with ambiguous boundaries.

In this article, we present an automated deep learning pipeline for deriving annotations of skeletal regions (legs, pelvis, lumbar/thoracic/cervical spine, ribcage, and arms/shoulders), along with adjacent internal organs (liver, spleen, urinary bladder, and kidneys) and the spinal canal contents (spinal cord and CSF) from b-value images and ADC maps. We use a weakly-supervised approach where soft-labels derived from a previously described atlas-based segmentation method, which generates a voxel-wise probability of each voxel belonging to a specific region. Although these labels are less precise than manual annotations, they enable the training of a significantly faster weakly-supervised deep learning model that leverages an extensive multi-centre dataset of WB-DWI scans from patients diagnosed and treated for APC or MM. As a result, the trained model can quickly generate reproducible probability maps for localising and delineating body regions from WB-DWI.

Our innovative weakly-supervised deep learning model could address the limitations of existing algorithms by: (i) reducing registration errors caused by initial spatial alignment between target and atlas images, (ii) decreasing the computational time required for whole-body registration, (iii) eliminating the need for time-consuming manual annotations by radiologists during model development, (iv) offering soft-boundary maps that capture anatomical structures with unclear boundaries due to disease spread, and the low resolution and geometric distortion of DWI.

## 2. Methods

### 2.1 Patient Population

To train the weakly-supervised deep learning model, we used three anonymised and retrospective WB-DWI datasets. **Dataset A** consisted of 217 patients diagnosed with APC, and **Dataset B** comprised 127 patients with confirmed MM (**Table 1**). These datasets were obtained from three different imaging centres, each varying in experience with WB-DWI protocols for oncological investigations. Finally, **Dataset C** consisted of 15 patients (13 male, 2 female) with confirmed MM with diffuse pattern disease, with images captured at a single centre. These images were manually contoured and defined our cohort for atlas-based segmentation.

In **Dataset A**, each patient underwent both pre- and post-treatment scanning, typically performed three months after treatment initiation to assess response or disease progression. These scans were acquired at one of three imaging centres, with 188, 22, and 7 patients per centre, respectively. For patients with confirmed MM (**Dataset B**), baseline scans were obtained at one of two imaging centres, with 113 and 14 patients per centre. Additionally, 12.6% of MM patients had a post-treatment scan, which was included in the dataset alongside the baseline scans from all patients.

The total number of scans from male and female patients was 497 and 80, respectively. Patients from each dataset were divided into training, validation, and test cohorts. Two experienced radiologists [*NT and CM blinded for review*] in functional cancer imaging and WB-DWI selected 15 patients with APC and 15 patients with MM from **Datasets A** and **B**, respectively, for the test cohort. The remaining patients (314) were split (80:20) into training and validation cohorts, resulting in 251/425 patients/scans for the training set and 63/107 patients/scans for the validation set.

Patient selection for the test cohort was stratified to capture a range of disease characteristics, including volume (e.g., low vs. high), pattern (e.g., focal vs. diffuse), extent of skeletal involvement, and anatomical location. To evaluate model performance on an independent set, a radiologist and a research scientist [*RD and AC, blinded for review*], each with 6+ years of experience in WB-DWI, manually delineated six randomly selected axial slices per body region on the DWI sequence across all 30 test scans (~2000 slices total). Full-body delineation typically requires 5–6 hours per scan; therefore, this slice-based approach was used to ensure feasibility. Importantly, these test scans were not used for generating soft-labels using the atlas-based registration algorithm.

All data were fully anonymised, and the study was performed following the Declaration of Helsinki (2013). A local ethical committee waived the requirement of patient consent for the use of these retrospective datasets (Approval Ref: The Royal Marsden Hospital Service Evaluations 696/2018 and SE1218/2023).

**2.2 Whole-body MRI protocol**

WB-DWI scans were acquired between 2015 and 2023 using two (50/900 s/mm$^2$) or three (50/600/900 s/mm$^2$) b-value diffusion weighted images on a 1.5T scanner (MAGNETOM Aera/Avanto, Siemens Healthcare, Erlangen, Germany). The scans covered 5-7 stations from the skull base to mid-thigh (for APC) or from the skull vertex to the knees (for MM). Each station consisted of 40 slices with a slice thickness of 5-6 mm. DWI was acquired as axial 2D single shot echo-planar imaging with acceleration factor 2 (GRAPPA) and three/four scan trace encoding directions. Additionally, STIR fat suppression with an inversion time of 180ms was employed to enhance image quality and contrast. All MRI parameters for DWI sequence are reported in **Table 1** for each dataset involved in the study. The whole-body MRI protocol also provides complementary morphological and functional imaging through the acquisition of sagittal T1/T2 spine images and 3D T1w DIXON images [25]. This is a valid and widely accepted protocol, included in radiological guidelines and recommendations for reporting metastatic bone disease (MET-RADS-P [26]) and patients diagnosed with MM (MY-RADS-P [27]).

**Table 1**. Scanning protocol and MRI parameters for all WB-DWI datasets investigated in our study. Minimum and maximum values are displayed in parenthesis.

| | **APC - Dataset (A) (217 patients, 434 WB-DWI scans)** | | **MM - Dataset (B) (127 patients, 143 WB-DWI scans)** | | **MM Atlas - Dataset (C) (15 patients, 15 WB-DWI scans)** | |
|---|---|---|---|---|---|---|
| **Dataset Split** | Training: 160 patients, 320 scans<br><br>Validation: 42 patients, 84 scans<br><br>Test: 15 patients, 30 scans | | Training: 91 patients, 105 scans<br><br>Validation: 21 patients, 23 scans<br><br>Test: 15 patients, 15 scans | | Leave-one-out cross-validation (LOOCV) for registration algorithm,<br><br>Test dataset for deep-learning models | |
| **MR scanner** | 1.5T Siemens Aera | | 1.5T Siemens Aera | | 1.5T Siemens Aera | |
| **Sequence** | Diffusion-Weighted SS-EPI | | Diffusion-Weighted SS-EPI | | Diffusion-Weighted SS-EPI | |
| **Acquisition plane** | Axial | | Axial | | Axial | |
| **Breathing mode** | Free breathing | | Free breathing | | Free breathing | |
| **b-values** $[s/mm^2]$ | b50/b900 (N = 21) | b50/b600/b900 (N = 196) | b50/b900 (N = 25) | b50/b600/b900 (N = 102) | b50/b900 (N = 6) | b50/b600/b900 (N = 9) |
| **Number of averages per b-value** | (3,5) | [(2,2,4) - (3,6,6)] | (4,4) | (2,2,4) | (4,4) | (3,3,3) |
| **Reconstructed resolution** $[mm^2]$ | [1.56x1.56 - 3.12x3.12] | | [1.54x1.54 - 1.68x1.68] | | [1.54x1.54 - 1.68x1.68] | |
| **Slice thickness [mm]** | [5 - 6] | | 5 | | 5 | |
| **Repetition time [ms]** | [5490 - 12700] | | [6150 - 14500] | | [6150 - 14500] | |
| **Echo time [ms]** | [60 - 79] | | [66.4 - 69.6] | | [64 - 69.9] | |
| **Inversion time (STIR fat suppression) [ms]** | 180 | | 180 | | 180 | |
| **Flip angle [°]** | 90 | | 90 | | 90 | |
| **Reconstructed matrix [mm]** | [98x128 - 256x256] | | [208x256 - 224x280] | | [208x256 - 224x280] | |
| **Receive bandwidth [Hz/Px]** | [1955 - 2330] | | [1984 - 2330] | | [1953 - 2330] | |

### 2.3 AI-model for body regions localisation and delineation

An overview of the proposed pipeline for our weakly-supervised deep learning model is illustrated in **Figure 1**.

#### 2.3.1 Soft-label annotation using atlas-based registration

To generate soft-labels for deep-learning model training, we employed a previously described atlas-based technique [28]. For all patients in our atlas cohort (**Dataset C**), a consultant radiologist [*AD blinded for review*] with 5+ years of experience in functional cancer imaging, manually delineated the whole skeleton and internal organs on the DWI sequence across all 15 scans. They assigned labels to seven different skeletal regions (legs, pelvis, lumbar/thoracic/cervical spine, ribcage, and arms/shoulders) and four internal organs (liver, spleen, urinary bladder, and kidneys). This cohort only included MM patients with diffuse disease for the following reasons: (i) MM scans have generally the widest coverage (from skull vertex to knees) within our centre, and (ii) patients diagnosed with diffuse disease pattern exhibits elevated skeleton diffusion signal allowing the bone to be more easily identified.

For segmentation of each new patient in the training/validation cohort of **Datasets A** and **B** (“target data”), we firstly registered the calculated ADC maps from each of the fifteen delineated patients in the atlas cohort (“atlas data”). Two AI-based models were employed for optimising the initial alignment between the atlas and target data, as previously described [28], [29]: (i) a body region classifier that automatically predicts the position of axial images into one of six potential body regions (legs, pelvis, lumbar/thoracic/cervical spine, and head), and (ii) a model that automatically delineates the spinal cord and surrounding CSF using a trained 2D U-Net model. After applying this initial alignment, registration was fine-tuned using affine followed by non-linear diffeomorphic demons transformation [30]. We utilised a mean squared error (MSE) optimisation between the target and atlas ADC data. For affine registration, we employed an Amoeba optimiser (simplex delta = 0.005) with shrink factors (4, 2, 1), smoothing sigmas (4, 2, 0), and a maximum of 200 iterations. For deformable demons registration, Gradient Descent optimisation was applied (learning rate = 2) with non-rigid shrink factors (4, 2, 1), smoothing widths (4, 2, 0), and a convergence window size of 20. The convergence tolerance for both methods was set to $10^{-6}$.

The derived transforms were then applied to all the pre-countered body region masks in the atlas cohort. Transformed masks from the same region were combined using a weighted summation, where the weights are the reciprocal of the MSE values:

$$P_j = \frac{\sum_{i=1}^{15} \frac{1}{MSE_i} \cdot T_i(M_{ij})}{\sum_{i=1}^{15} \frac{1}{MSE_i}} \qquad (3)$$

Where:

- $P_j$ is a probability for each voxel that the voxel belongs to tissue class $j$.
- $T_i$ is the registration transform derived for atlas patient *i*.
- $M_{ij}$ denotes the (untransformed) mask for tissue class $j$ in each atlas patient *i*.
- $MSE_i$ is the mean squared error in ADC values between atlas patient *i* and the target new patient following registration.

The labels for the spinal cord and surrounding CSF were directly derived from the 2D U-Net model, resulting in 12 probability maps, as illustrated in **Figure 1**. Probability maps were stacked to generate the soft-labels in tensor form, each including 13 channels per WB-DWI scan, where the last channel was dedicated to representing the image background: $P_{background} = 1 - \sum_{j=1}^{12} P_j$, which is guaranteed to remain between 0 and 1.

### 2.3.2 Weakly-supervised model training

A 3D patch-based Residual U-Net model (Res U-Net) [31] was trained to generate skeleton, internal organ, and spinal canal probability maps, using soft-labels obtained from the soft-label annotation phase. The networks involved a 2-channel input: (i) the ADC map (no thresholding was applied to remove negative ADC calculations) and (ii) the estimated intercept (S0) image at b = 0 s/mm$^2$. The ADC map and S0 image were derived by fitting a monoexponentially decaying model to the diffusion data included in the training and validation cohorts [32], [33]. The multi-channel, SoftMax output of the model was the probability maps of seven skeleton regions, four internal organs, spinal canal, and in the last channel the image background. All images in the training and validation cohorts were interpolated to matrix = 256 x 256 and resolution = 1.6 x 1.6 mm. Input images were normalised using the following transformations:

$$scaled\ ADC\ map\ = ADC\ map\ /\ 3.5 \cdot 10^{-3}\ mm^2/s \quad (4)$$

$$scaled\ S0\ image = \log(S0\ image)\ /\ \max\ (\log\ (S0\ image)) \quad (5)$$

The Res U-Net symmetrical encoder-decoder architecture consisted of 5 convolutional layers, with the number of filters down-sampled or up-sampled by a factor of 2 at each layer, starting from 32 filters. Each layer employed skip connections between corresponding encoder and decoder paths. All convolutional layers used a 3x3 kernel size with a stride of 2, batch normalisation, a dropout rate of 0.2, and the ReLU activation function, except for the final output layer, which used a SoftMax activation function. Residual connections were implemented in each layer of the encoder and decoder paths to mitigate the vanishing gradient problem. We employed stochastic gradient descent optimization with hyperparameters: learning rate of 0.1, momentum of 0.9, and weight decay of $4 \cdot 10^{-5}$, aimed at minimizing a multi-class cross-entropy loss function. The training process lasted 300 epochs, using a batch size of 4 and patch size of 128x128x64. Patches were randomly sampled from the full volume during training to ensure representative coverage of anatomical structures. At inference, volumetric delineations were generated using a sliding window algorithm with 25% patch overlap to reduce edge artifacts and improve prediction consistency at low-resolution boundaries. All algorithms were implemented in Python (v.3.7) using PyTorch v.1.12.1 and MONAI v.0.9.1 toolboxes, running upon a Windows platform (v.10.0.19) accelerated by an NVIDIA RTX6000 GPU (Santa Clara, California, US). After training, the weakly-supervised Res U-Net model can automatically generate probability maps for body regions from a new WB-DWI scan using pre-processed ADC maps and S0 images, eliminating the need for computing atlas-based registration algorithms.

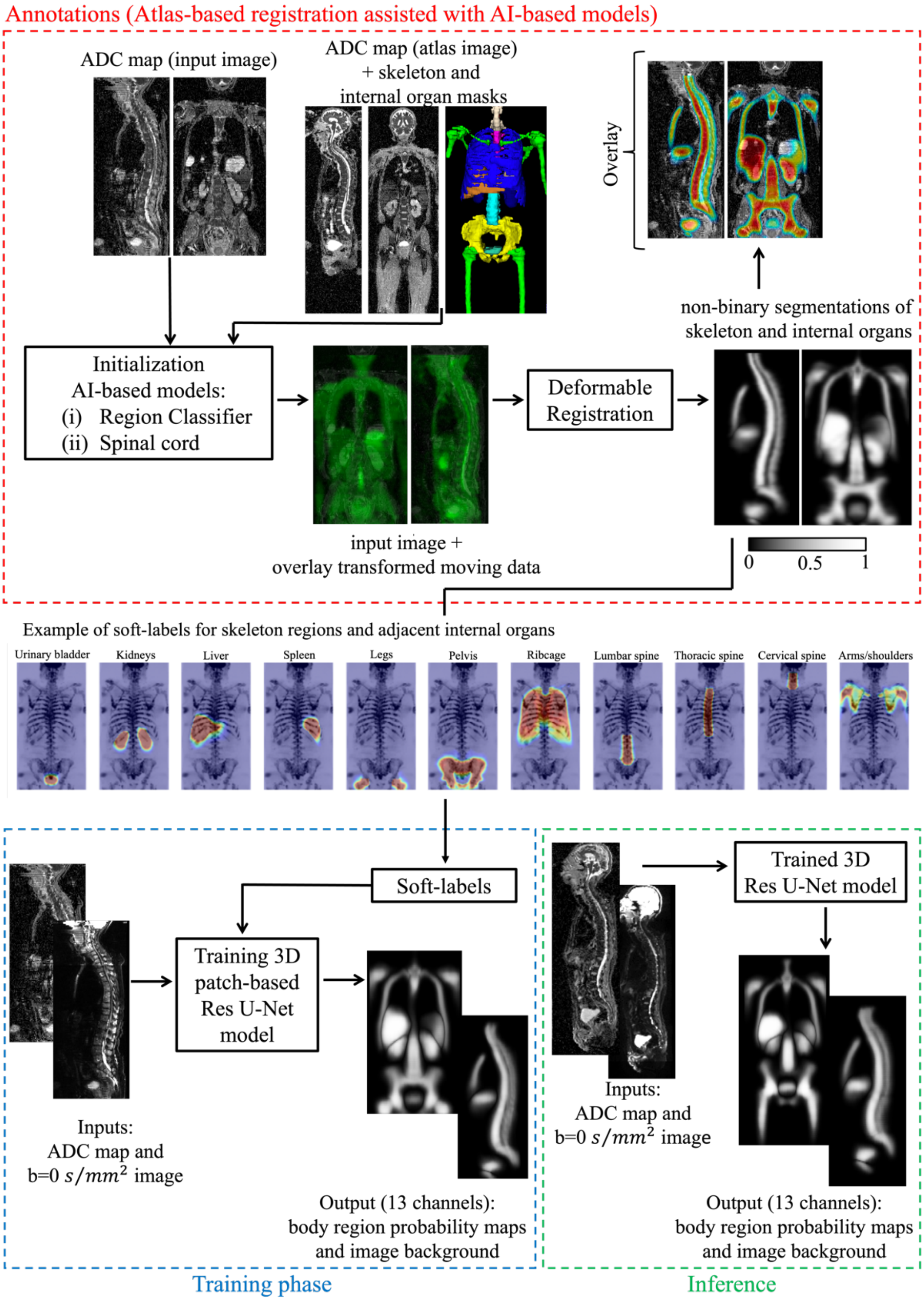

**Figure 1**. Automated end-to-end pipeline for developing a weakly-supervised deep learning model that automatically localises/delineates seven skeleton regions (legs, pelvis, ribcage, lumbar/thoracic/cervical spine, and arms/shoulders), four adjacent internal organs (liver, spleen, urinary bladder, and kidneys) and the spinal canal from WB-DWI scans. We employed an atlas-based registration assisted with two previously trained AI-based models for generating soft-labels (non-binary segmentations). Next, we used the soft-labels for training a faster 3D patch-based Res U-Net model with 2-channel input, the ADC map and b = 0 s/mm$^2$ image, and as output (13 channels) the body region probability maps alongside the image background. As a result, for a new input WB-DWI, the trained Res U-Net model can generate all body region probability maps without any registration step.

## 2.4 Evaluation Criteria

### 2.4.1 Probability Calibration

Probabilities generated by machine learning models must be calibrated so that outputs reflect true likelihoods [34]. We calibrated the output of our deep learning model using all patients in **Datasets C**: each channel of the SoftMax output of our model was partitioned into 20 equidistant intervals, $x$, with boundaries [0, 0.05, …, 0.95, 1]; for pixels that fell within each partition we then calculated the proportion occupied with that ground-truth label, $y$, from manual delineation. The generated curves then indicate whether the model output under- or over-estimates true label frequencies during segmentation and thus represents a valid probability distribution [35].

We used isotonic regression to calibrate the model outputs, which ensures a monotonic relationship between the predictor and target variables [36]. This technique is formulated as an optimisation problem, expressed as:

$$min \sum_{i=1}^{n}(y_i - f(x_i))^2 \text{ subject to } f(x_1) \leq f(x_2) \ldots \leq f(x_n) \qquad (6)$$

For our multi-class model, we employed a one-vs-all strategy [37]. Each channel was calibrated in turn, with the remaining channels being grouped together as the negative class. The success of the weakly-supervised model calibration was evaluated by calculating the log-loss between the true labels and calibrated probability maps derived from the isotonic regression fitting. A decrease in the log-loss indicates the success of the calibration process, ensuring true probabilities and mitigating under- and over-predictions by the trained Res U-Net model.

### 2.4.2 Quantitative analysis

We assessed the accuracy of our calibrated deep-learning Res U-Net model to expert annotations using two test sets: (i) an independent cohort of 30 patients (15 APC and 15 MM) from **Datasets A** and **B**, in which six randomly selected axial slices per body region were manually annotated (~2000 slices total); and (ii) the atlas patient cohort (**Dataset C**), which was also used to derive the soft-labels for model training. However, the ADC maps and S0 images from the atlas dataset were not used directly for training. Therefore, we argue that the results from both cohorts would form a valid test of our model's performance.

To compare our approach with that of conventional atlas-based segmentation on test cohort (ii), we employed a leave-one-out-cross-validation (LOOCV) strategy. For each of the 15 patients, we derived (1) an atlas-based segmentation using the method described in section 2.3.1 using the remaining 14 patients, and (2) a segmentation as estimated using our proposed deep-learning model. For test cohort (i), all 15 atlas images were used to generate atlas-derived maps.

A threshold of 0.45 (empirically chosen to balance sensitivity and specificity) was applied to convert the body region probability maps from both methods into binary masks. We then computed the average surface distance between manual and automated delineations. Additionally, we transferred the automated and manual body region masks onto the ADC map to calculate ADC statistics (median) and report relative differences. For overlap-based metrics, we computed the dice similarity coefficient (DSC) for atlas-based delineations. However, for the Res U-Net model, we employed a continuous (soft) DSC that directly compares the model's calibrated probability maps with binary reference masks [38]. This avoids applying an arbitrary threshold on inherently probabilistic outputs and aligns with our aim to retain uncertainty information. Significant differences between the results derived from both automated methods were assessed using a Wilcoxon signed-rank test ($p < 0.05$ indicating significance), with the best method selected based on the highest DSC and the lowest computational time.

### 2.4.3 Semi-quantitative analysis

Two consultant radiologists [*AD and RD*], blinded to each other's ratings, completed a semi-quantitative analysis to evaluate the accuracy of our segmentation model in our 30 test-patient cohort (15 APC and 15 MM). They assessed segmentation quality by superimposing estimated probability maps for all body regions onto the ADC maps and high b-value images ($b = 900$ s/mm$^2$). A qualitative assessment was made using a 4-point Likert scale: 1 = segmentation failure, 2 = suboptimal performance, 3 = good accuracy, and 4 = excellent accuracy. Criteria considered for each WB-DWI scan assessment were: (i) the degree to which the location of the body region probability maps matched the actual patient anatomy, further verified by reviewing T1/T2 spine and whole-body DIXON T1w images; (ii) the success of the derived delineations in covering the volume of the skeleton and internal organs visible on the WB-DWI scan; (iii) how well the automated delineations separated different anatomical structures (e.g., ensuring predicted kidney masks did not overlap with actual liver boundaries). Inter-reader agreement was quantified using both percentage agreement and Gwet's AC1 [39], which provides stable estimates in the presence of unbalanced rating distributions. Agreement levels were classified according to the categories proposed by Landis and Koch [40]. Finally, this expert assessment would provide insights into whether the automated tool is sufficiently developed to serve as preliminary step in isolating the whole bone marrow background, facilitating the development of signal-based AI tools for automated bone lesion detection/delineation from WB-DWI scans.

#### 2.4.4 Benchmarking with open-source model

We benchmarked our weakly-supervised, calibrated Res U-Net model against the open-source *TotalSegmentator-MRI* nnU-Net (version 2.7.0) [21]. Whole-body b = 50 s/mm² DWI images were used as input, as this sequence best matched the modalities used to train the open-source model. We computed dice scores between automated masks and manual annotations for the urinary bladder, kidneys, liver, spleen, long bones, pelvis, spine, and spinal canal across all test cases (Section 2.4.2). The ribcage was excluded, as it is not supported by the *TotalSegmentator-MRI* model. Dice scores for each region were then compared across the three automated methods: atlas-based registration, *TotalSegmentator-MRI*, and our Res U-Net model. Statistical differences were assessed using the Wilcoxon signed-rank test ($p < 0.05$).

## 3. Results

### 3.1 Probability Calibration

In **Figure 2**, we demonstrated the reliability curves for each body region in **Dataset C**. These curves revealed that the weakly-supervised Res U-Net model tended to overestimate or underestimate predictions compared to the true probabilities. In the uncalibrated models for binary classification, data points consistently fell below or above the ideal calibration line. However, following calibration using the isotonic regression technique, we observed a notable decrease in log-loss (on average -15%) compared to the uncalibrated models, with data points moving closer to the ideal calibration line.

### 3.2 Quantitative analysis

**Figure 3** shows the overlap-based metrics, along with average surface distances and relative median ADC differences, computed from both manual and automated masks following calibration for each body region. The weakly-supervised Res U-Net model outperformed the atlas-based registration algorithm, with average DSC values for all body regions increasing by more than 8%, demonstrating statistically significant differences between the methods. The computational time required to generate probability maps from the model averaged 25 seconds, compared to the 5 minutes needed for the atlas-based registration algorithm on the same hardware (2.4 GHz Quad-Core Intel Core i5).

Across test cases in **Datasets A** and **B**, our deep learning model achieved a DSC of 0.73 for the long bones, 0.78 for the pelvis, and 0.82 for the spine. However, lower performance was observed for ribcage delineations, with a DSC below 0.5. For internal organs, dice scores exceeded 0.77, except for the spleen, which showed a score of 0.71. The average surface distance was below 2mm, and relative median ADC differences were below 13% for both skeletal and internal organ masks.

Consistent results were observed for test cases in **Dataset C**. The model achieved the following results for skeleton delineations: a DSC of 0.67 with an average surface distance between automated and manual expert-defined delineations of less than 2mm. Significant improvements compared to the atlas-based registration algorithm were observed in the pelvis and spine, where we obtained a DSC of 0.73 and 0.84. For long bones these overlap-based were slightly lower, averaging just below 0.7. However, performance decreased for the ribcage, with a DSC of 0.43. In terms of internal organs, our model achieved a DSC of 0.86. The spleen yielded the lowest score among soft tissues, with an average surface distance exceeding 3mm. We observed a statistically significant reduction in relative median ADC differences for whole skeleton delineations compared to the atlas-based registration algorithm. The relative median ADC differences for the skeleton and internal organs between automated and manual expert-defined delineations were below 10.5% and 4%, respectively, except for the bladder, which exceeded 14%. Spinal canal delineations showed a DSC of 0.86, consistent with results from the 2D U-Net model. **Figures 4** and **5** present manual and derived delineations of the skeleton, internal organs, and spinal canal for three randomly selected test cases (**Dataset C**), confirming the weakly-supervised deep learning model's ability to localise and delineate body regions from WB-DWI scans.

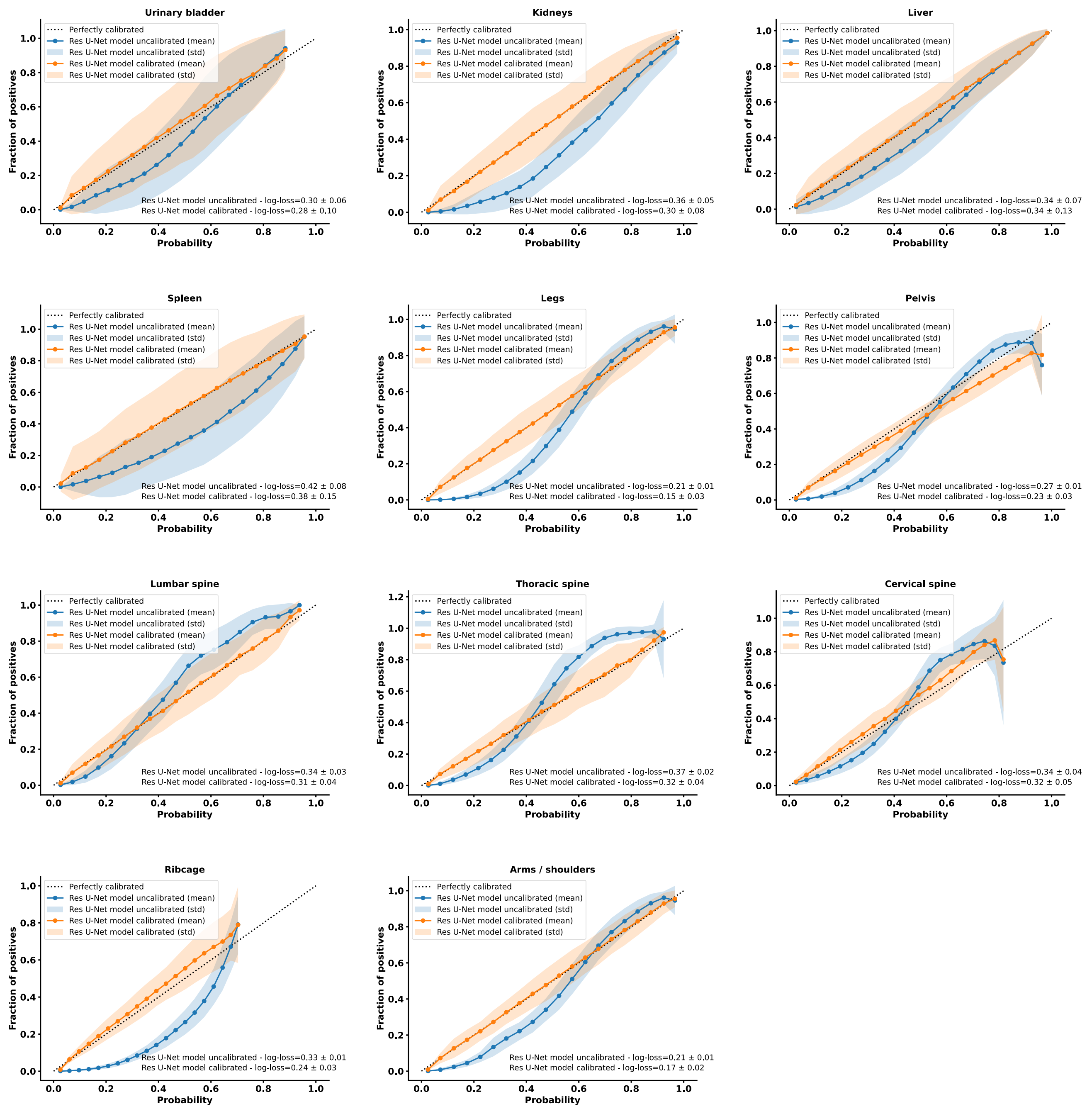

**Figure 2**. Reliability curves used to evaluate the calibration of the weakly-supervised Res U-Net model. To perform this analysis, we used all patients included in the atlas cohort (**Dataset C**), with a leave-one-out testing approach. Confidence intervals (mean and standard deviation) were calculated for each data point in the plots. We generated 11 reliability curves, assuming models trained for a binary classification task (one-vs-all strategy). Importantly, predictions (all voxels) from the SoftMax activation function (uncalibrated model) tended to overestimate or underestimate true probabilities, as evidenced by data points falling below or above the y=x line. In contrast, models calibrated using the isotonic regression technique demonstrated a reduced log-loss between true labels and predicted probabilities, with data points aligning closer to the ideal calibration curves. This calibration process ensures that model predictions accurately reflect true probabilities, thereby mitigating the risks associated with under or over predictions.

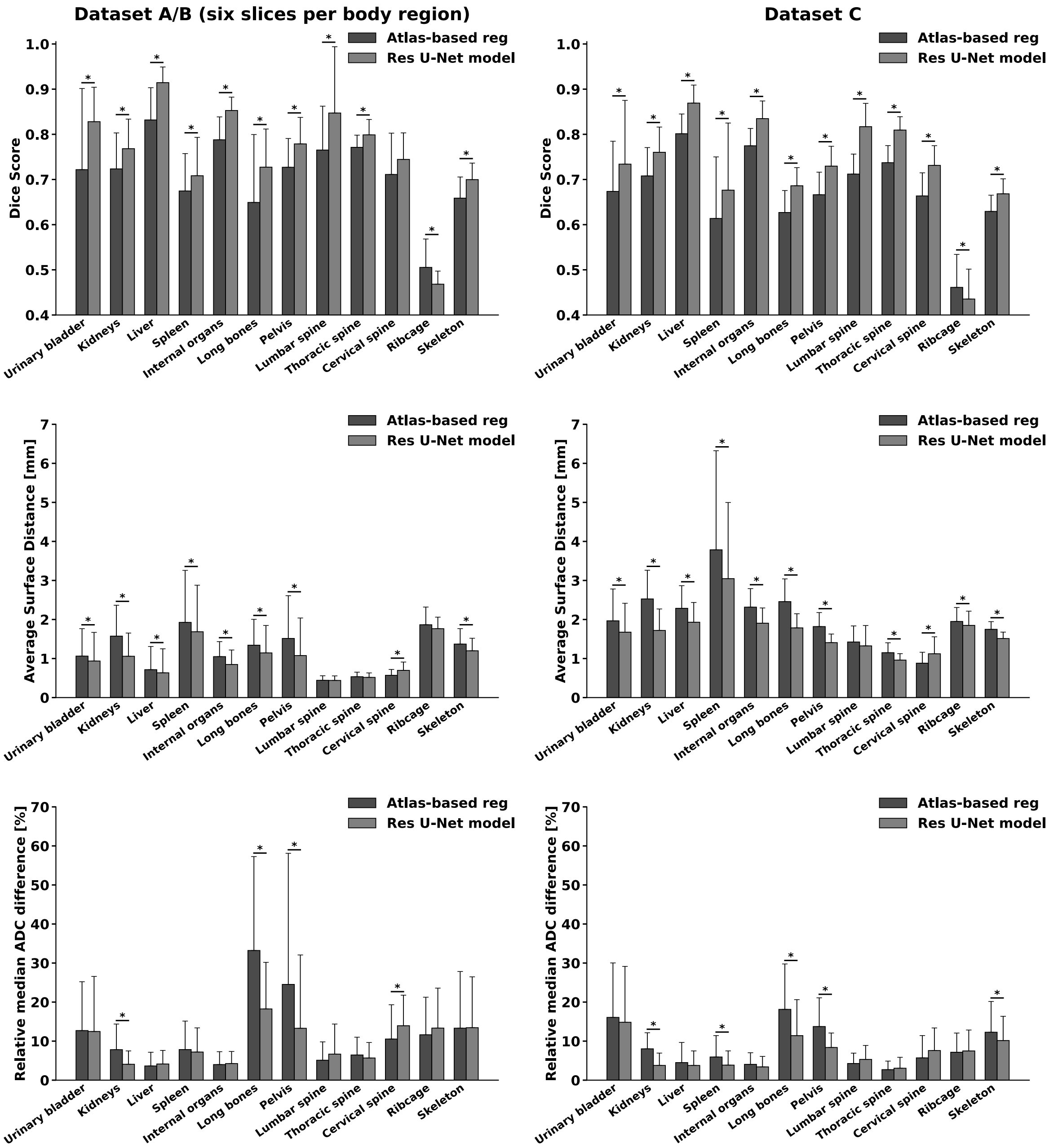

**Figure 3**. Quantitative metrics for evaluating the performance of the calibrated weakly-supervised 3D Res U-Net model in localising and delineating body regions from WB-DWI scans, compared to expert-defined delineations. Additionally, the same metrics were derived using the atlas-based registration algorithm (baseline, through LOOCV for **Dataset C**). Overlap-based metrics showed a significant improvement for the deep-learning model over the registration algorithm across all body regions, with at least an 8% increase in DSC. Moreover, the model generated outputs 12 times faster than the registration algorithm (25 seconds versus 5 minutes). Our weakly-supervised model also showed lower relative median ADC differences, with statistically significant improvements observed for whole skeleton delineations. Mask for long bones consisted of legs and arms/shoulders. Mask for skeleton included long bones, pelvis, lumbar/thoracic/cervical spine, and ribcage. Mask for internal organs included urinary bladder, kidneys, liver and spleen.

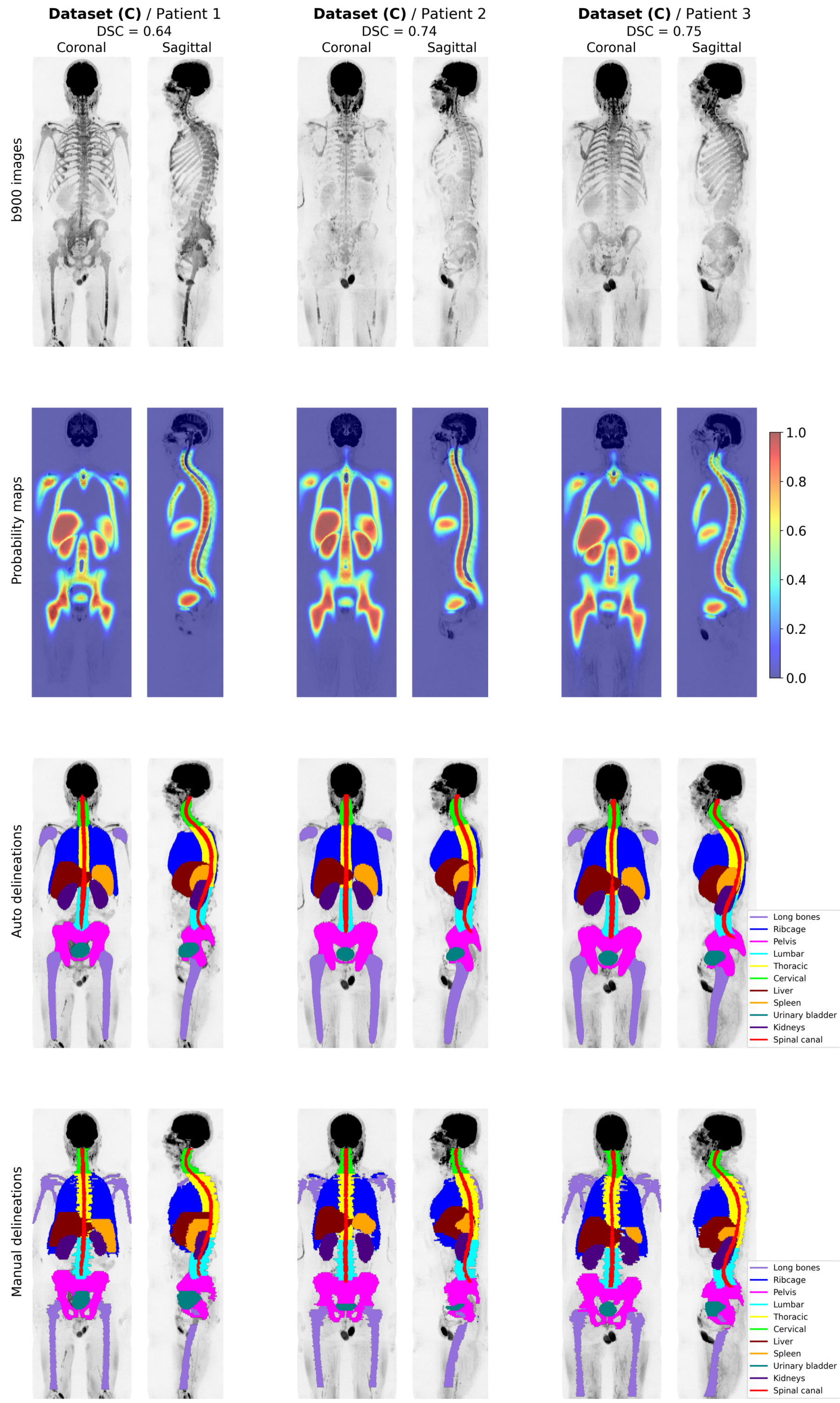

**Figure 4**. Coronal and sagittal views of Maximum Intensity Projection (MIP) of b900 images for three randomly selected test cases from **Dataset C** (top row). The second to fourth rows show body region probability maps from the weakly-supervised, calibrated Res U-Net model (where values close to "1" indicate a high probability that voxels belong to one of the body regions), body region masks (obtained by thresholding the probability maps), and manually expert-defined labels, all superimposed on the MIP images. Automated localisation and delineations of all body regions showed promising agreement with the actual patient anatomy, with dice scores of 0.64, 0.74, and 0.75 for the three cases when comparing manual and automated masks. However, lower performance is observed in delineating the upper long bones. Mask for long bones consisted of legs and arms/shoulders.

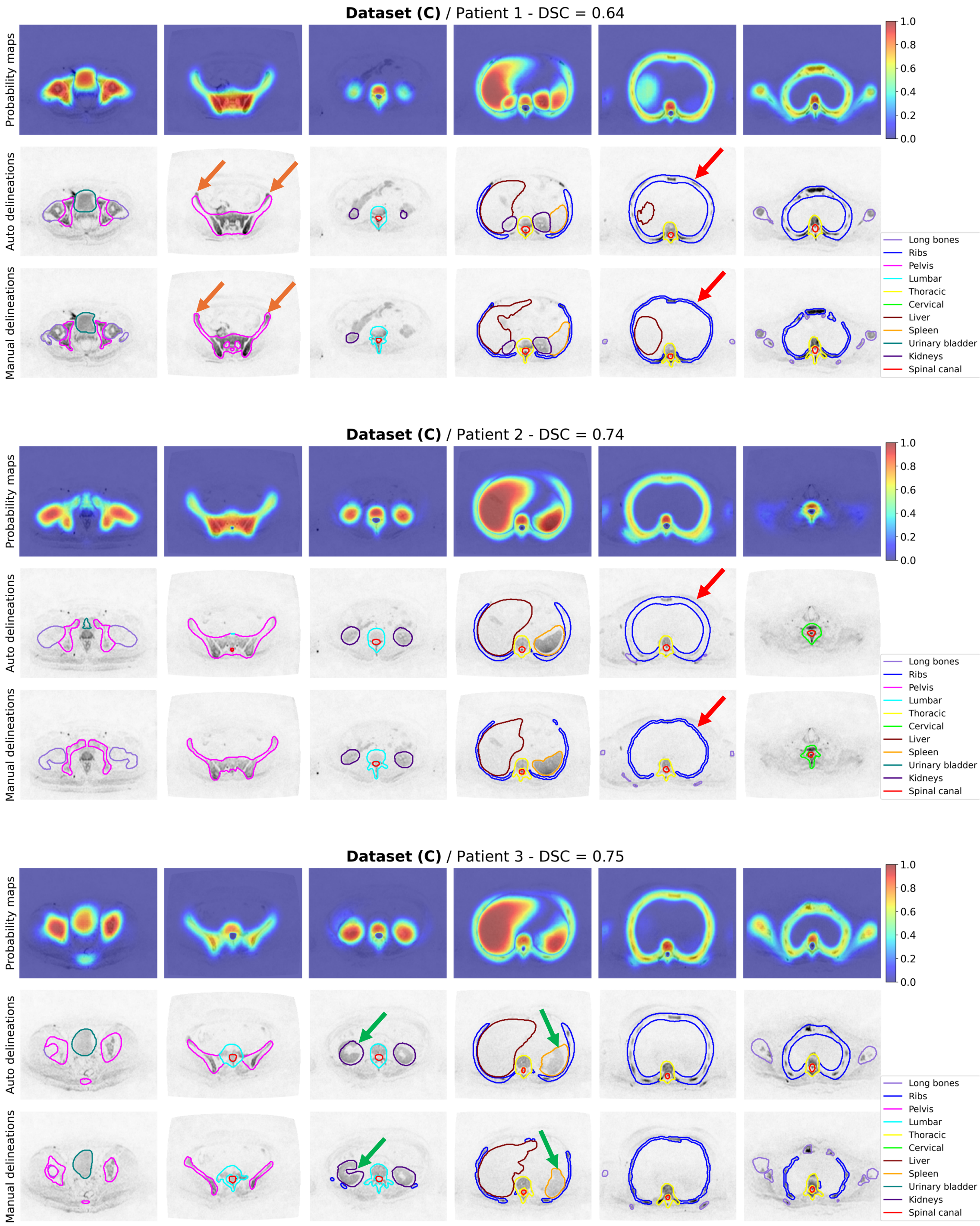

**Figure 5**. Axial b900 images with superimposed manual and automated delineations of the skeleton, adjacent internal organs, and spinal canal for three randomly selected test cases from **Dataset C**. The weakly-supervised, calibrated Res U-Net model demonstrated high accuracy in localising different anatomical structures, as shown by the substantial agreement between the generated probability maps (where values close to "1" indicate a high probability that voxels belong to one of the body regions) and actual patient anatomy. Dice scores of 0.64, 0.74, and 0.75 were achieved for the three cases when comparing manual and automated masks. However, the model presented limitations in capturing complex soft-tissue organ shapes (green arrows), showed misalignment in the pelvic bone delineation for *"Patient 1"* **(**orange arrows**)**, and generated over-segmentation of the ribcage in all cases (red arrows). Mask for long bones consisted of legs and arms/shoulders.

### 3.3 Semi-quantitative analysis

Radiologists rated the calibrated probability maps for all test cases as either "**good**" or "**excellent**", with 80% (24/30) inter-reader agreement across these two categories based on overall scores (substantial agreement; Gwet's AC1=0.72). Agreement by anatomical region was 63.3% (19/30) for the skeleton and 53.3% (16/30) for internal organs, reflecting fair agreement levels (Gwet's AC1=0.38 and 0.27, respectively); see **Supplementary Material Tables 1** and **2**.

**Figures 6** and **7** show three randomly selected patients with APC and three with MM. The generated body region probability maps aligned well with the patients' anatomy, showing no signs of overfitting. Both the skeleton and internal organs were accurately localised within the scans, with probabilities exceeding 0.4, while the image background displayed values close to 0, as expected. Importantly, our weakly-supervised Res U-Net model generated reproducible skeletal probability maps in test patients with heterogenous metastatic and active myeloma bone disease, reflecting the stratified selection criteria used in our test cohort.

General observations provided by the radiologists included: (i) excellent segmentations of the spinal canal; (ii) a tendency to under-segment the inferior pubic rami, but no false positives from including pelvic organs/vessels; (iii) no false positives from thigh vessels; (iv) no false positives from axillary lymph nodes. This latter point is particularly encouraging, given that lymph nodes were not delineated as part of this model development.

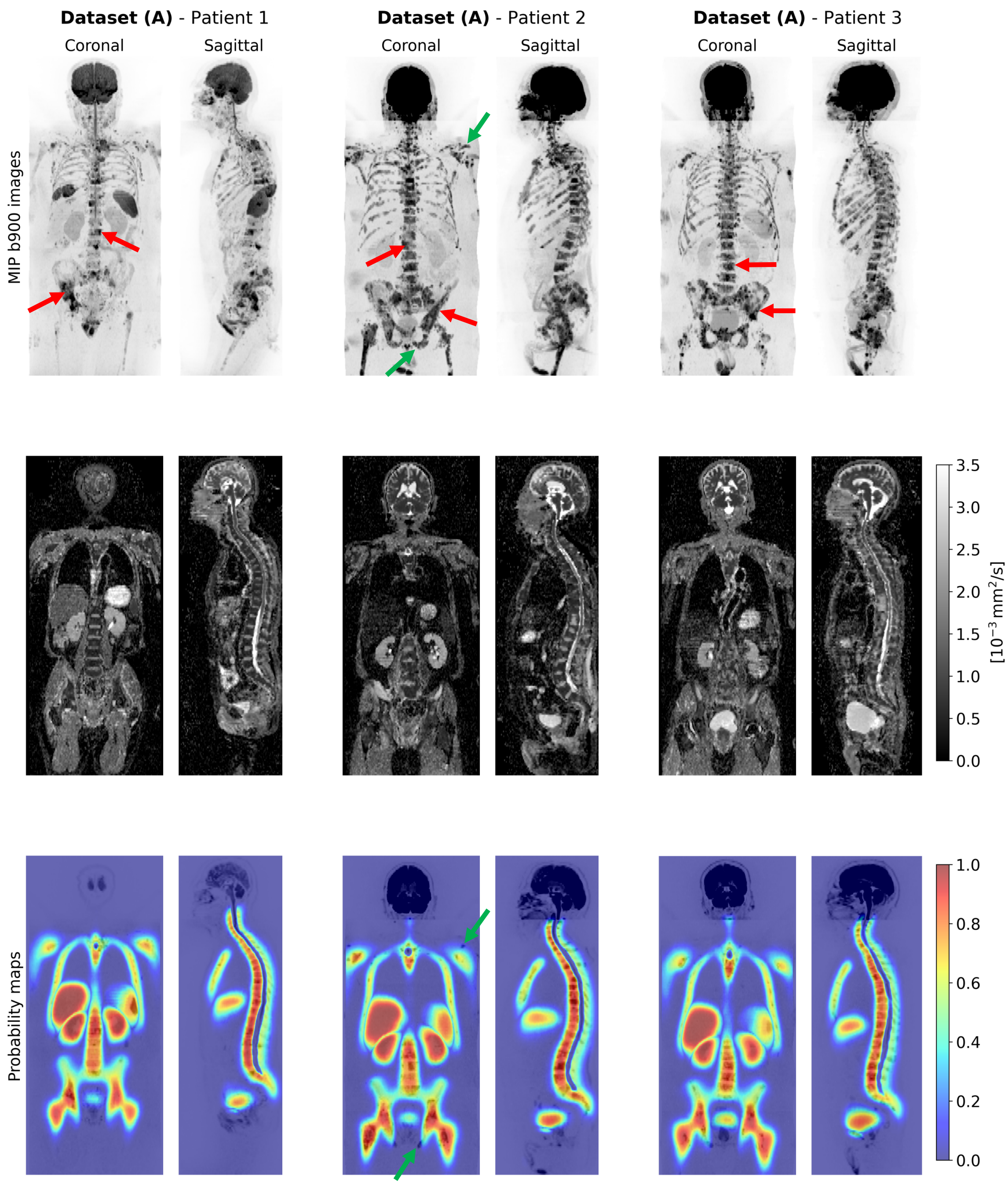

**Figure 6**. Coronal and sagittal views of Maximum Intensity Projection (MIP) of b900 images and ADC maps for three randomly selected APC patients from the test cohort (**Dataset A**) with confirmed diffuse metastatic bone disease. These lesions (red arrows) appear as darker spots with high signal intensity on b900 images and low ADC values (typically $>5{\cdot}10^{-3}$mm/s$^2$) compared to the normal bone marrow background. Superimposed on these MIP images are body region probability maps generated by the weakly-supervised, calibrated Res U-Net model. The derived probability maps (where values close to "1" indicate a high probability that voxels belong to one of the body regions) accurately matched patient anatomy, ensuring correct positional alignment, volume coverage, and promising boundary delineation between skeletal regions and adjacent internal organs. These findings were supported by expert semi-quantitative analysis, which rated the automated full-body probability maps for these patients as "excellent" (score of 4), except for the skeleton map in "*Patient 2*", which was rated as "good" (score of 3) due to lower accuracy in localising the top of the shoulders and pubic rami (green arrows).

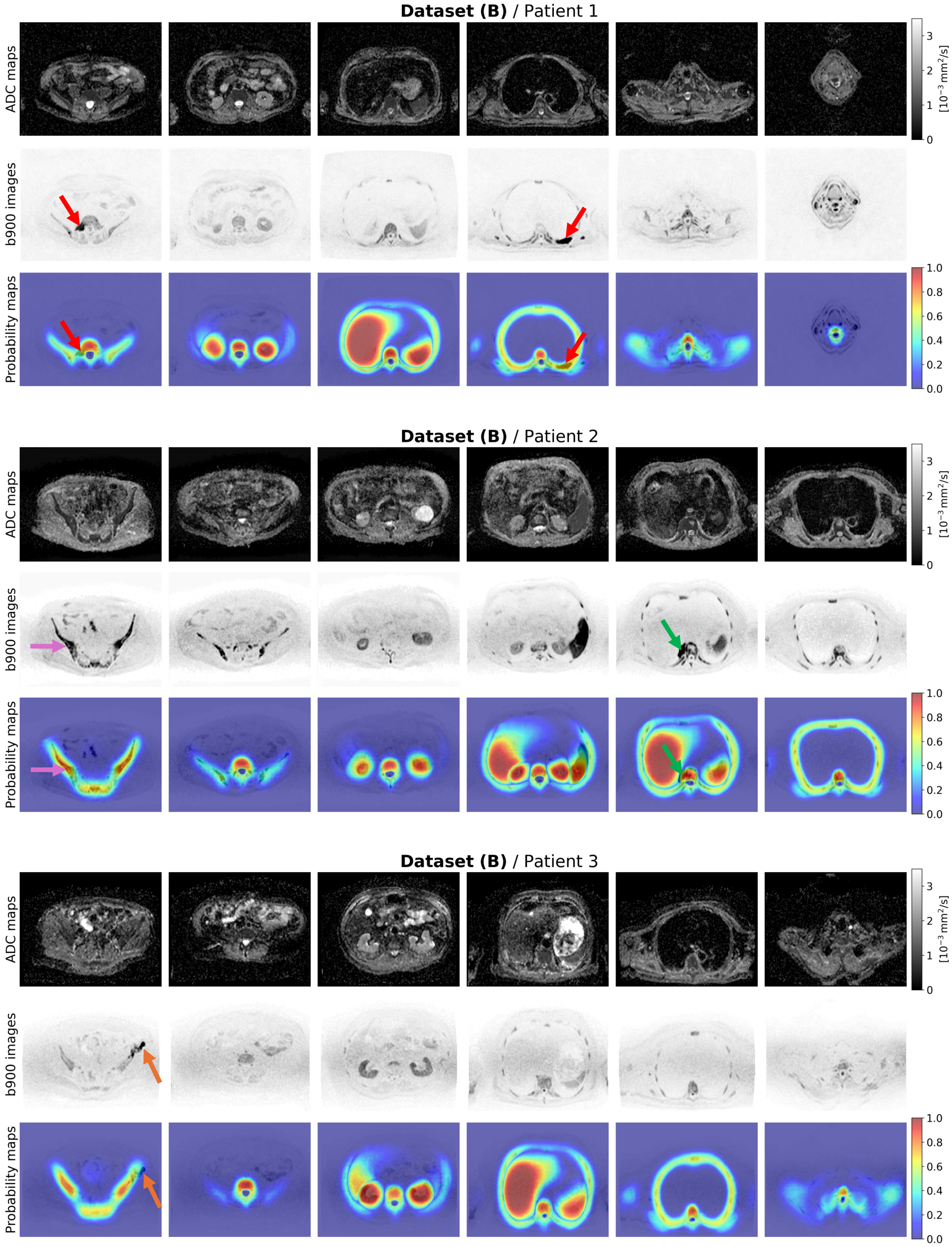

**Figure 7**. Axial ADC maps and b900 images for three patients diagnosed with MM from the test cohort (**Dataset B**), presenting different disease patterns: focal bone lesions (red arrows), diffuse myeloma involvement (purple arrows), and paramedullary disease (green arrows). Body region probability maps derived from the weakly-supervised, calibrated Res U-Net model were superimposed on the b900 images. These maps demonstrated accurate positional alignment, volume coverage, and clear separation between skeletal regions and adjacent internal organs compared to actual patient anatomy. Expert evaluation rated them as "excellent" (score of 4). However, the skeleton map for "*Patient 3"* was rated as "good" (score of 3) due to reduced accuracy in delineating the anterior iliac (orange arrows).

### 3.4 Benchmarking with open-source model

Our weakly-supervised, calibrated Res U-Net model outperformed *TotalSegmentator-MRI* in skeleton (excluding ribcage) delineation, achieving a DSC of 0.78 vs. 0.64 ($p<0.05$) for cases in **Datasets A** and **B**, and 0.76 vs. 0.60 ($p<0.05$) in **Dataset C**, see **Tables 2** and **3**. The spine showed the largest performance gap, with scores of 0.82 vs. 0.49 ($p<0.05$) in **Datasets A** and **B**, and 0.84 vs. 0.57 in **Dataset C**. Spinal canal delineation also demonstrated a significant difference in **Datasets A** and **B** (0.79 vs. 0.60, $p <0.05$), while no significant difference was observed in **Dataset C** (0.86 vs. 0.81, $p\geq0.05$). Although *TotalSegmentator-MRI* achieved slightly higher dice scores for internal organ delineation in **Dataset C** (0.86 vs. 0.83), the difference was not statistically significant ($p\geq0.05$). The spleen showed the most notable performance gap between the models (0.78 vs. 0.68, $p<0.05$). For internal organs in **Datasets A** and **B**, both models performed similarly, with dice scores of 0.85. Finally, our solution generated delineations in 25 seconds per WB-DWI scan, compared to 12 minutes for the open-source model, with both tested on a CPU (Apple M3 Max, 16-core).

**Table 2**. Dice scores computed between expert-defined body region annotations and the outputs of three automated methods across test cases from **Datasets A** and **B** (only six slices per body region per scan were annotated): (i) atlas-based registration algorithm, (ii) weakly-supervised, calibrated Res U-Net model, and (iii) the open-source *TotalSegmentator-MRI* model. The highest scores with statistically significant differences (Wilcoxon signed-rank test, $p < 0.05$) are shown in bold*.

| Datasets A/B | Atlas-based registration | Res U-Net model (ours) | *TotalSegmentator-MRI* |
|---|---|---|---|
| Urinary bladder | 0.72 ± 0.18 | 0.83 ± 0.08 | 0.85 ± 0.06 |
| Kidneys | 0.72 ± 0.08 | 0.77 ± 0.07 | **0.83 ± 0.06*** |
| Liver | 0.83 ± 0.07 | **0.91 ± 0.03*** | 0.86 ± 0.05 |
| Spleen | 0.67 ± 0.08 | 0.71 ± 0.09 | **0.82 ± 0.05*** |
| ***Internal organs*** | 0.79 ± 0.05 | 0.85 ± 0.03 | 0.85 ± 0.03 |
| Long bones | 0.65 ± 0.15 | 0.73 ± 0.08 | 0.74 ± 0.10 |
| Pelvis | 0.73 ± 0.06 | **0.78 ± 0.06*** | 0.67 ± 0.10 |
| Spine | 0.76 ± 0.04 | **0.82 ± 0.05*** | 0.49 ± 0.08 |
| ***Skeleton*** (excl. ribcage) | 0.72 ± 0.05 | **0.78 ± 0.04*** | 0.64 ± 0.07 |
| Spinal canal | - | **0.79 ± 0.05*** | 0.60 ± 0.05 |
| ***All body regions*** (excl. ribcage) | 0.76 ± 0.04 | **0.83 ± 0.03*** | 0.77 ± 0.03 |

**Table 3**. Dice scores computed between expert-defined body region annotations and the outputs of three automated methods across test cases from **Datasets C:** (i) atlas-based registration algorithm (through LOOCV), (ii) weakly-supervised, calibrated Res U-Net model, and (iii) the open-source *TotalSegmentator-MRI* model. The highest scores with statistically significant differences (Wilcoxon signed-rank test, $p < 0.05$) are shown in bold*.

| Datasets C | Atlas-based registration | Res U-Net model (ours) | *TotalSegmentator-MRI* |
|---|---|---|---|
| Urinary bladder | 0.67 ± 0.11 | 0.73 ± 0.14 | 0.81 ± 0.06 |
| Kidneys | 0.71 ± 0.06 | 0.76 ± 0.06 | **0.82 ± 0.02*** |
| Liver | 0.80 ± 0.04 | 0.87 ± 0.04 | 0.88 ± 0.05 |
| Spleen | 0.61 ± 0.14 | 0.68 ± 0.15 | **0.78 ± 0.06*** |
| ***Internal organs*** | 0.77 ± 0.04 | 0.83 ± 0.04 | 0.86 ± 0.03 |
| Long bones | 0.63 ± 0.05 | 0.69 ± 0.04 | 0.70 ± 0.08 |
| Pelvis | 0.67 ± 0.05 | 0.73 ± 0.04 | 0.73 ± 0.06 |
| Spine | 0.76 ± 0.03 | **0.84 ± 0.03*** | 0.57 ± 0.04 |
| ***Skeleton*** (excl. ribcage) | 0.67 ± 0.03 | **0.76 ± 0.03*** | 0.67 ± 0.03 |
| Spinal canal | - | 0.86 ± 0.05 | 0.81 ± 0.05 |
| ***All body regions*** (excl. ribcage) | 0.73 ± 0.02 | **0.80 ± 0.02*** | 0.76 ± 0.02 |

## 4. Discussion

We developed and validated a weakly-supervised deep learning model based on the 3D Res U-Net architecture to automatically localise and delineate seven skeleton regions (legs, pelvis, lumbar/thoracic/cervical spine, ribcage, and arms/shoulders), four internal organs (liver, spleen, urinary bladder, and kidneys), and the spinal canal (spinal cord with surrounding CSF) from WB-DWI scans. We trained a model using an heterogenous WB-DWI dataset comprising 532 scans from patients with APC or MM, without requiring resources for laborious manual delineations of body regions. Instead, we automated the annotation phase using a validated atlas-based registration algorithm. Manual annotations of body regions can take up to 5-6 hours per WB-DWI scan; however, with our novel methodology, we generated annotations (soft-labels) for 532 scans (~127,500 images) within ~48 hours (~325 seconds/scan).

The quantitative analysis demonstrated that our weakly-supervised Res U-Net model improved accuracy in localising and delineating skeleton and internal organs on test datasets compared to the atlas-based registration algorithm (baseline). Across all body regions, the model achieved mean DSC of 0.80 compared to 0.69 for the baseline ($p<0.05$). For skeleton delineation alone, the DSC was 0.67 vs. 0.63 ($p<0.05$), and for internal organs, 0.83 vs. 0.77 ($p<0.05$). In contrast to atlas-based registration algorithms, which rely on predefined templates, our deep learning model trained on heterogeneous datasets captures diverse features and patterns, enhancing spatial dependencies and structural information present in WB-DWI scans, thereby leading to improved generalisation on unseen datasets. However, lower performance was observed for ribcage (DSC <0.5) and spleen delineations (average surface distance values exceeding 5mm). This may be due to geometric fidelity issues in the ribcage region from DWI sequence and variability in spleen position between patients due to respiratory motion during free-breathing acquisition [24].

Spinal canal delineation accuracy was consistent with a previously validated 2D U-Net model, achieving a DSC of 0.86. Our previous study proven that accurate spinal canal delineations can significantly improve the inter- and intra-patient signal normalisation on WB-DWI [29]. This approach can enhance assessment of bone disease from WB-DWI by improving lesion localisation, diameter measurement, and "at-a-glance" evaluation of post-treatment changes by standardising the display of high-b-value images.

Fast computation and accurate localisation of body region probability maps are key for facilitating clinical deployment of automated tools for disease staging and treatment response assessment. The computational time required to generate probability maps from the weakly-supervised model averaged 25 seconds, compared to the 5 minutes needed for the atlas-based registration algorithm on the same hardware. Furthermore, relative median ADC difference between manual and automated skeleton segmentations was 10% (95% CI: 6.59-13.7%), which is closely comparable with previously reported inter-reader repeatability of ADC measurements (6%; 95% CI: 2.70–11.5%), based on manual experts' delineations of bone lesions in APC patients [41], [42].

The semi-quantitative analysis provided further evidence of the model's generalisability on unseen WB-DWI data from patients with APC and MM. The model showed no signs of overfitting and generated accurate delineations of multiple organs even in patients with extensive disease burden. Two experienced radiologists rated cases as either "good" or "excellent" (on a 4-point Likert scale: "1: failed", "2: suboptimal", "3: good", "4: excellent").

We also compared the performance our solution with the open-source *TotalSegmentator-MRI* model. Overall, our calibrated Res U-Net achieved higher dice scores (0.8 vs. 0.76, $p<0.05$), with the most significant improvement observed in the skeletal delineation (0.76 vs. 0.67, $p<0.05$). These results are encouraging given the key advantages of our model: (i) it is designed to generate soft-boundary probability maps, improving delineation of anatomical structures with unclear boundaries due to disease spread, as well as the low resolution and geometric distortion of DWI. This should minimise the risk of fully excluding disease involved areas from the delineated volume; (ii) it achieves fast inference times, approx. 25 seconds per WB-DWI scan compared to 12 minutes for *TotalSegmentator-MRI*, when both were tested on a CPU. This reduces dependence on dedicated GPUs, which are less available in hospitals for on-premises AI model deployment.

However, there are limitations to our current solution. Firstly, although we plan to include additional WB-DWI datasets with (semi)manual annotations in future work, the current study reports results from a test cohort in which only six axial slices per body region were manually annotated. This sampling strategy may not fully capture anatomical variability between patients. In the second test cohort, full-body annotations were available; however, the same images were also used during atlas-based registration to generate soft-labels, potentially introducing bias.

Secondly, the semi-quantitative analysis involved only two readers. Including more readers with varying levels of experience in interpreting WB-DWI could provide a more robust reference for assessing the accuracy and reproducibility of the generated body region probability maps.

Thirdly, the model was trained and tested using images acquired from a single MRI vendor. However, the automated annotation phase could aid in the model's re-training or application of transfer learning techniques. This would involve using WB-DWI scans acquired from different scanner manufacturers and/or leveraging novel AI-based reconstruction methods [43]. Such strategies aim to enhance the model's ability to generalise to unseen data without the need for manual delineation of skeleton regions and internal organs.

Fourthly, our model demonstrated lower performance (DSC=0.76, excluding the ribcage) compared to previously published algorithms developed using anatomical whole-body MRI for bone marrow segmentation in patients with malignant bone disease, as well as to reported inter-reader agreement levels (DSC=0.88) [19]. Ceranka et al. [15] developed a multi-atlas segmentation algorithm from T1w MRI, reporting a dice score of 0.88 in a LOOCV study with 10 APC patients. However, their method excluded the ribcage and upper limbs and required 180 minutes per scan to generate a mask. Bauer et al. [20] trained a nnU-Net on 210 coronal T1w MRI scans to delineate the spine, pelvis, humeri, and femora, also achieving a dice score of 0.88 in 21 test patients with myeloma-related pathologies. However, their approach relied on T1w coronal MRI, which is not included in the MY-RADS guidelines for myeloma assessment and required 18 minutes per scan for inference. Instead, Kedzierska et al. [44] developed a nnU-Net for WB-DWI using multi-centre datasets. However, their model focused only on pelvic bone marrow, reporting dice scores of 0.92 and 0.85 for the hip and sacral bones, respectively. In our study, the inter-reader agreement for ADC measurements of the pelvic bone (71 $x10^{-6}$ mm²/s, mean absolute difference) aligned with their results (65-85 $x10^{-6}$ mm²/s), supporting reliable ADC quantification despite lower segmentation overlap.

These lower overlap-based metrics may be explained by two reasons: (i) our deep learning model was trained using soft-labels generated from atlas-based registration, rather than high-quality manual full-body annotations, to facilitate training on a high-volume, multi-centre data; (ii) Standardisation of the WB-DWI protocol is not a trivial task due to the significant number of parameters that can vary across scanners, leading to variability in the quality of the images acquired at different sites, including variations in geometric distortion and SNR. This makes alignment with anatomical MRI sequences difficult, and further challenging organ delineations.

In conclusion, our novel weakly-supervised deep learning model, based on the 3D Res U-Net architecture, can swiftly generate accurate and reproducible probability maps for localising skeletal regions, internal organs, and the spinal canal from WB-DWI scans. This will facilitate the development of automated tools for delineating suspected bone disease, based on WB-DWI signal intensity. This in turn could enable the extraction of quantitative imaging biomarkers for staging and treatment response assessment. Prospective evaluation of our tool in a multi-centre setting is currently ongoing, as part of a national trial assessing the effectiveness of these tools for automatically delineating metastatic bone disease or multifocal myeloma lesions from WB-DWI in patients with APC and MM.

**Suppl. Material Table 1**. Semi-quantitative analysis of derived skeleton and internal organ probability maps from the weakly-supervised, calibrated Res U-Net model, evaluated by two experienced radiologists using scans acquired from patients with metastatic bone disease (test cohort, **Dataset A**). Probability maps were rated on a 4-point Likert scale: 1 = failed, 2 = suboptimal, 3 = good, 4 = excellent.

| APC test cohort (Dataset A) | Skeletal probability map | | Internal organ probability map | | Overall score | | Radiologists' comments |
|---|---|---|---|---|---|---|---|
| | R1 | R2 | R1 | R2 | R1 | R2 | |
| **001** | 3 | 4 | 4 | 3 | 4 | 4 | |
| **002** | 4 | 3 | 3 | 4 | 4 | 3 | Part of the scapula missing and the pelvis includes some soft tissue |
| **003** | 4 | 4 | 4 | 4 | 4 | 4 | |
| **004** | 3 | 3 | 3 | 4 | 3 | 3 | Part of the spleen included in the ribcage. Also top of the shoulders not included in skeleton. A part of a retroperitoneal lymph nodes included under low probability spine |
| **005** | 4 | 4 | 4 | 3 | 4 | 4 | Top of the left shoulder not included in skeleton |
| **006** | 4 | 4 | 4 | 4 | 4 | 4 | Part of spleen included under ribcage |
| **007** | 4 | 3 | 4 | 3 | 4 | 3 | |
| **008** | 4 | 3 | 4 | 4 | 4 | 4 | Part of the scapula (inferior angle) missing and the pelvis includes some soft tissue |
| **009** | 4 | 4 | 4 | 3 | 4 | 4 | |
| **010** | 4 | 3 | 4 | 4 | 4 | 4 | Very good negotiation between retroperitoneal lymph nodes and spine, and between spleen and ribcage |
| **011** | 3 | 4 | 4 | 4 | 4 | 4 | Tops of shoulder, inferior pubic rami, and right pubic bone disease under very low probability, [however high volume of disease so these minor areas are not significant] |
| **012** | 3 | 3 | 3 | 4 | 3 | 3 | Part of the spleen included in the ribcage, top of the shoulders not included in skeleton, and misalignment in the pelvic bone |
| **013** | 4 | 4 | 4 | 4 | 4 | 4 | |
| **014** | 4 | 4 | 4 | 4 | 4 | 4 | |
| **015** | 4 | 3 | 4 | 3 | 4 | 3 | |

**Suppl. Material Table 2.** Semi-quantitative analysis of derived skeleton and internal organ probability maps from the weakly-supervised, calibrated Res U-Net model, evaluated by two experienced radiologists using scans acquired from MM patients (test cohort, **Dataset B**). Probability maps were scored on a 4-point Likert scale: 1 = failed, 2 = suboptimal, 3 = good, 4 = excellent.

| **MM test cohort (Dataset B)** | **Skeleton probability map** | | **Internal organ probability map** | | **Overall score** | | **Radiologists' comments** |
|---|---|---|---|---|---|---|---|
| | R1 | R2 | R1 | R2 | R1 | R2 | |
| **001** | 4 | 4 | 4 | 3 | 4 | 4 | |
| **002** | 3 | 3 | 4 | 3 | 4 | 3 | Expansile disease in the sternum |
| **003** | 3 | 4 | 4 | 3 | 4 | 3 | |
| **004** | 4 | 4 | 4 | 4 | 4 | 4 | |
| **005** | 3 | 4 | 4 | 4 | 4 | 4 | Missed inferior pubic rami, top of acromioclavicular joints, and distal right femur |
| **006** | 4 | 4 | 3 | 4 | 4 | 4 | right kidney partially included in liver |
| **007** | 4 | 4 | 4 | 4 | 4 | 4 | Very difficult scan. Excellent negotiation of large artefacts from breast implants |
| **008** | 4 | 4 | 4 | 4 | 4 | 4 | Some tips of transverse processes and very lateral end of right clavicle under very low probability, [however, in the face of such high-volume disease these are insignificant] |
| **009** | 4 | 4 | 4 | 4 | 4 | 4 | |
| **010** | 4 | 4 | 4 | 4 | 4 | 4 | Excellent demarcation between artefact and right 1st rib |
| **011** | 3 | 4 | 4 | 4 | 4 | 4 | Very top of the shoulders under very low probability |
| **012** | 4 | 4 | 4 | 3 | 4 | 4 | Post edge of paramedullary disease sternum on the very low probability values - 0.05 |
| **013** | 4 | 4 | 4 | 4 | 4 | 4 | Part of paramedullary disease head of lower right rib on low probability values - 0.2 |
| **014** | 4 | 4 | 4 | 3 | 4 | 3 | Two lesions top of shoulders - included!<br>Misalignment for kidneys and spleen |
| **015** | 3 | 4 | 4 | 4 | 4 | 4 | Part of left anterior iliac lesion under very low probability. Same with left lateral and bilateral mid clavicles |

**Author Contributions**

Conceptualization, A.C., M.D.B. and D.-M.K.; methodology, A.C. and M.D.B.; software, A.C. and R.H.; validation, A.C., R.H., M.D.B., N.T., C.M., A.D. and D.-M.K.; formal analysis, A.C.; resources, R.D., C.M. and N.T.; data curation, A.C. and R.H.; writing-original draft preparation, A.C.; writing-review and editing, A.C., R.H., A.R., R.D., A.D., C.M., N.T., M.D.B. and D.-M.K.; supervision, M.D.B. and D.-M.K.; project administration, A.R.; funding acquisition, M.D.B. and D.-M.K. All authors have read and agreed to the published version of the manuscript.

- **Funding**

This project is funded by the NIHR Invention for Innovation award (Advanced computer diagnostics for whole body magnetic resonance imaging to improve management of patients with metastatic bone cancer II-LA-0216-20007).

- **Informed Consent Statement**

Patient consent was waived; study was conducted with retrospective data only and researchers only had access to de-identified data.

- **Data Availability Statement**

Data can be shared upon request from the authors. However, this is subject to the establishment of an appropriate data-sharing agreement, given the sensitive nature of patient information involved.

- **Acknowledgments**

This study represents independent research and we acknowledge support from the National Institute for Health and Care Research (NIHR) Biomedical Research Centre at The Royal Marsden NHS Foundation Trust and The Institute of Cancer Research, London, and by the Royal Marsden Cancer Charity and Cancer Research UK (CRUK) National Cancer Imaging Trials Accelerator (NCITA), and is funded by the NIHR Invention for Innovation award (Advanced computer diagnostics for whole body magnetic resonance imaging to improve management of patients with metastatic bone cancer II-LA-0216-20007). The views expressed are those of the author(s) and not necessarily those of the NIHR or the Department of Health and Social Care. This work uses data provided by patients and collected by the NHS as part of their care and support.
The authors would also like to acknowledge Mint Medical®.

- **Conflicts of interest**

A.C., R.H., and M.D.B. have submitted a patent to the UK Intellectual Property Office regarding the work described in this manuscript.